\documentclass[conference]{IEEEtran}
\IEEEoverridecommandlockouts
\usepackage{cite}
\usepackage{fancyhdr}
\usepackage{amsmath,amssymb,amsfonts}
\usepackage{algorithmic}
\usepackage{graphicx}
\usepackage{textcomp}
\usepackage{xcolor}
\def\BibTeX{{\rm B\kern-.05em{\sc i\kern-.025em b}\kern-.08em
    T\kern-.1667em\lower.7ex\hbox{E}\kern-.125emX}}
\fancypagestyle{FirstPage}{
\lfoot{979-8-3503-7565-7/24/\$31.00 \copyright2024 IEEE}
}
\begin{document}

\title{MultiMAE-DER: Multimodal Masked Autoencoder for Dynamic Emotion Recognition
}

\author{\IEEEauthorblockN{Peihao Xiang}
\IEEEauthorblockA{\textit{Department of Electrical and Computer Engineering} \\
\textit{Florida International University}\\
Miami, USA \\
pxian001@fiu.edu}
\and
\IEEEauthorblockN{Chaohao Lin}
\IEEEauthorblockA{\textit{Department of Electrical and Computer Engineering} \\
\textit{Florida International University}\\
Miami, USA \\
clin027@fiu.edu}
\and
\IEEEauthorblockN{Kaida Wu}
\IEEEauthorblockA{\textit{Department of Electrical and Computer Engineering} \\
\textit{Florida International University}\\
Miami, USA \\
kwu020@fiu.edu}
\and
\IEEEauthorblockN{Ou Bai}
\IEEEauthorblockA{\textit{Department of Electrical and Computer Engineering} \\
\textit{Florida International University}\\
Miami, USA \\
obai@fiu.edu}
}

\maketitle

\begin{abstract}
 This paper presents a novel approach to processing multimodal data for dynamic emotion recognition, named as the Multimodal Masked Autoencoder for Dynamic Emotion Recognition (MultiMAE-DER). The MultiMAE-DER leverages the closely correlated representation information within spatio-temporal sequences across visual and audio modalities. By utilizing a pre-trained masked autoencoder model, the MultiMAE-DER is accomplished through simple, straightforward fine-tuning. The performance of the MultiMAE-DER is enhanced by optimizing six fusion strategies for multimodal input sequences. These strategies address dynamic feature correlations within cross-domain data across spatial, temporal, and spatio-temporal sequences. In comparison to state-of-the-art multimodal supervised learning models for dynamic emotion recognition, MultiMAE-DER enhances the weighted average recall (WAR) by 4.41\% on the RAVDESS dataset and by 2.06\% on the CREMA-D. Furthermore, when compared with the state-of-the-art model of multimodal self-supervised learning, MultiMAE-DER achieves a 1.86\% higher WAR on the IEMOCAP dataset.
\end{abstract}


\section{Introduction}
\thispagestyle{FirstPage}
With the development of deep learning technology, convolutional neural networks (CNN) have shown excellent results in static emotion recognition, especially in facial expression recognition (FER) tasks, as seen in models like RTCNN \cite{c1}, DeepEmotion \cite{c2}, and PAtt-Lite \cite{c3}. Dynamic emotion recognition (DER) is one of the important fields of affective computing. Psychologists and neuroscientists have been exploring this field for a long time, while constructing a series of quantitative rules for recognizing emotional characteristics. 

\begin{figure}[htbp]
\centerline{\includegraphics[scale=0.12]{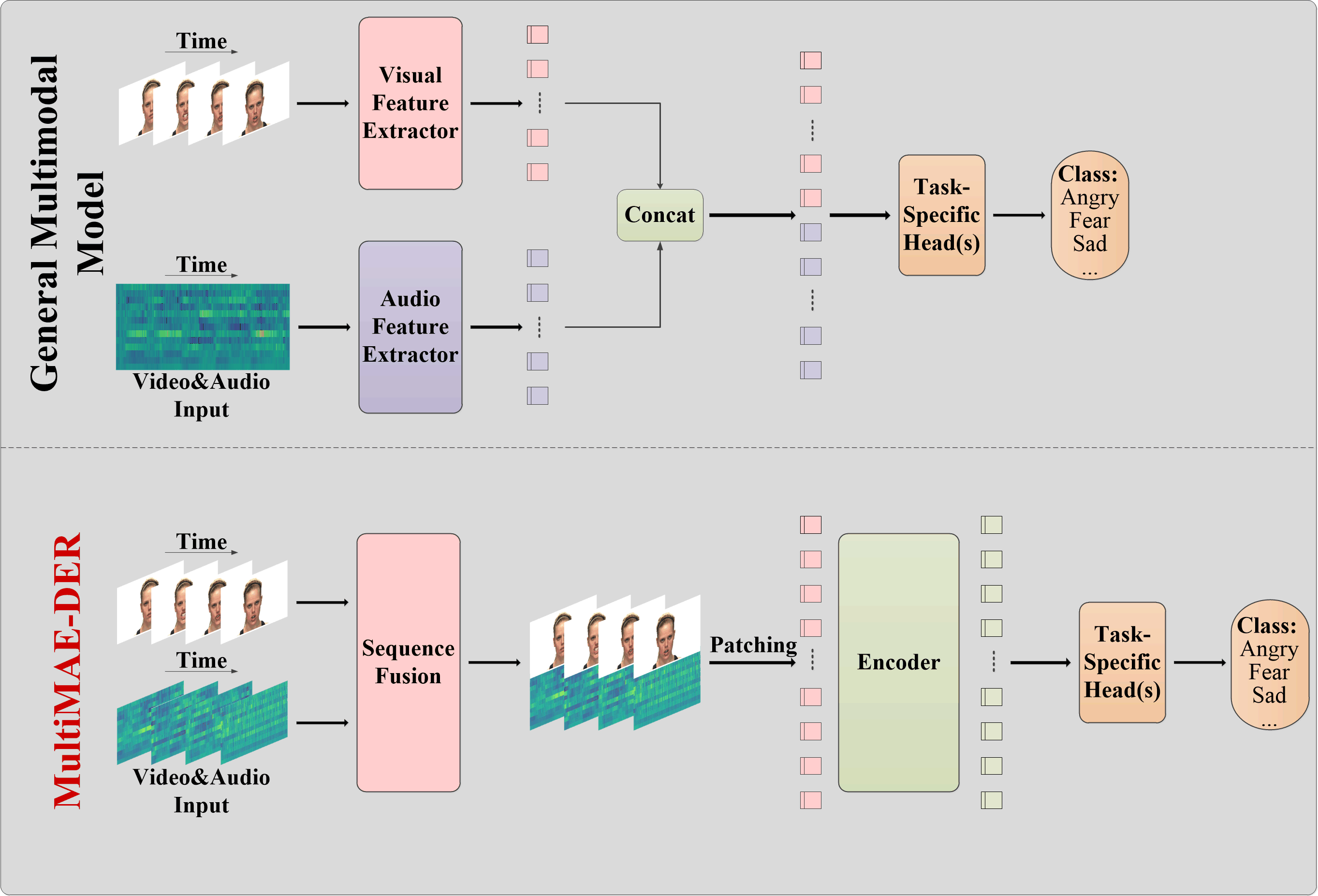}}
\caption{General Multimodal Model vs. MultiMAE-DER. The uniqueness of our approach lies in the capability to extract features from cross-domain data using only a single encoder, eliminating the need for targeted feature extraction for different modalities.}
\label{fig}
\end{figure}

In DER, a new architecture has emerged by combining CNN with recurrent neural networks (RNN). For example, ResNet-18+LSTM \cite{c4} and C3D+LSTM \cite{c5} have been developed to construct dynamic emotion recognition models. More importantly, with the advent of Transformers \cite{c6}, there has been further exploration into the global context correlation. This structure is well-suited for capturing the spatial and temporal context features in dynamic emotion. Examples of such models include Former-DFER \cite{c7} and STT-DFER \cite{c8}.

However, CNN, RNN, or Transformers \cite{c6}, requires a large amount of labeled data for supervised learning to build high-performance models. This issue was only alleviated with the advent of self-supervised learning, especially when applied to natural language processing (NLP), where the BERT \cite{c9} model demonstrated performance beyond supervised learning models.

In the field of computer vision, the introduction of Masked Autoencoder (MAE) strongly indicated that self-supervised learning would shine in the field of vision such as ImageMAE \cite{c10}, VideoMAE \cite{c11}, and AudioMAE \cite{c12}. On the other hand, while single-modal (visual or audio) inputs have shown good results in dynamic emotion recognition, as shown in models like HuBERT \cite{c13} and MAE-DFER \cite{c14}, their performance has not yet reached the level of real human emotion recognition. Therefore, multimodal (Visual-Audio) input has become a desirable development trend, as demonstrated by models like AV-LSTM \cite{c15} and MSAF \cite{c16}.

The aim of this study is to explore a new framework for dynamic emotion recognition model. The proposed approach is a straightforward inheritance and extension of VideoMAE \cite{c11}, extending the originally single-modal visual input into multimodal input encompassing both visual and audio elements. Simultaneously, the self-supervised learning pre-trained Video Masked Autoencoder model is used to process visual-audio sequence strategies. This extended model is named MultiMAE-DER, standing for Multimodal Masked Autoencoder for Dynamic Emotion Recognition. The MultiMAE-DER is optimized by investigating the performance of six visual-audio sequence strategies.

\section{RELATED WORK}

\subsection{Dynamic Emotion Recognition}

Dynamic emotion recognition is a significant challenge in the field of affective computing that requires the utilization of multimodal feature extraction, context semantic analysis, and causal reasoning techniques. In recent years, researchers have addressed the feature extraction task from multimodal using transfer learning by combining pre-trained models for video facial expression feature extraction and audio tone emotion feature extraction, such as MSAF \cite{c16} and CFN-SR \cite{c17}. However, the above-mentioned model architecture fails to establish cross-domain associative features between visual and audio data, leading to the loss of highly correlated spatio-temporal feature dimensions in the emotion representation information. Therefore, we propose a MultiMAE-DER framework, a unified feature extraction pre-trained model that will be employed for visual and audio multimodal data to find the associated features in visual-audio spatio-temporal emotion representation information.

\subsection{Masked Autoencoder}

Masked Autoencoder is a variant of denoising autoencoder that consists of an encoder and a decoder. In ImageMAE \cite{c10}, that is designed as an asymmetric structure to reconstruct normalized pixels. VideoMAE \cite{c11} follows the design approach of ImageMAE \cite{c10} but extends the 2D image input to 3D video input and incorporates a joint spatio-temporal self-attention mechanism to replace the vanilla Vision Transformer \cite{c18} spatial self-attention mechanism.

For downstream tasks, the decoder of the masked autoencoder is discarded, and only the pre-trained encoder is used for supervised learning and fine-tuning on the downstream task dataset. As a leader in self-supervised learning, our proposal will continue to inherit this characteristic and utilize the pre-trained encoder model to fine-tune multimodal dynamic emotion data, aiming for improved emotion recognition performance.

\subsection{Vision Transformer}
Vision Transformer (ViT)\cite{c18} is a variant of the Transformer \cite{c6} applied in the field of computer vision. The principle is to treat a pixel block (patch) as an individual ``word" to emulate the token sequence input of Transformer \cite{c6}, allowing for the serialization of image processing. Furthermore, ViT \cite{c18} serves as the backbone network for VideoMAE \cite{c11}, which is used to associate the correlation of spatio-temporal context in video data.

For MultiMAE-DER, ViT \cite{c18} will continue to be used as the backbone network for the encoder. Simultaneously, MultiMAE-DER will inherit the joint spatio-temporal self-attention mechanism from VideoMAE \cite{c11} to replace the spatial self-attention mechanism in vanilla Vision Transformer \cite{c18}. This adaptation aims to analyze the contextual emotion representation information in visual-audio data.

\begin{figure}[thpb]
  \centering
  \includegraphics[scale=0.5]{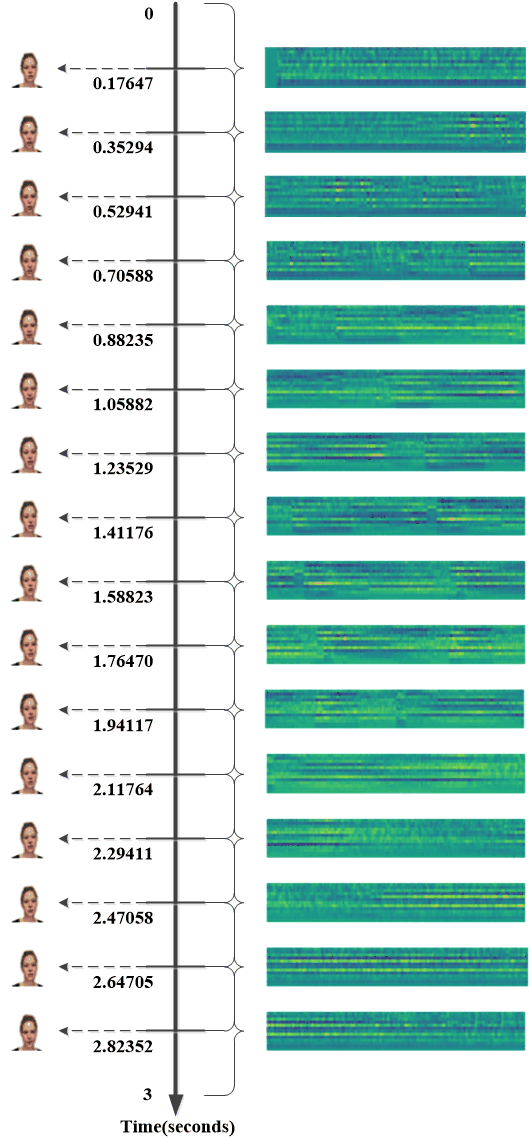}
  \caption{Time sequence of video frames and corresponding audio spectrograms.}
  \label{figurelabel}
\end{figure}

\begin{figure}[thpb]
  \centering
  \includegraphics[scale=0.14]{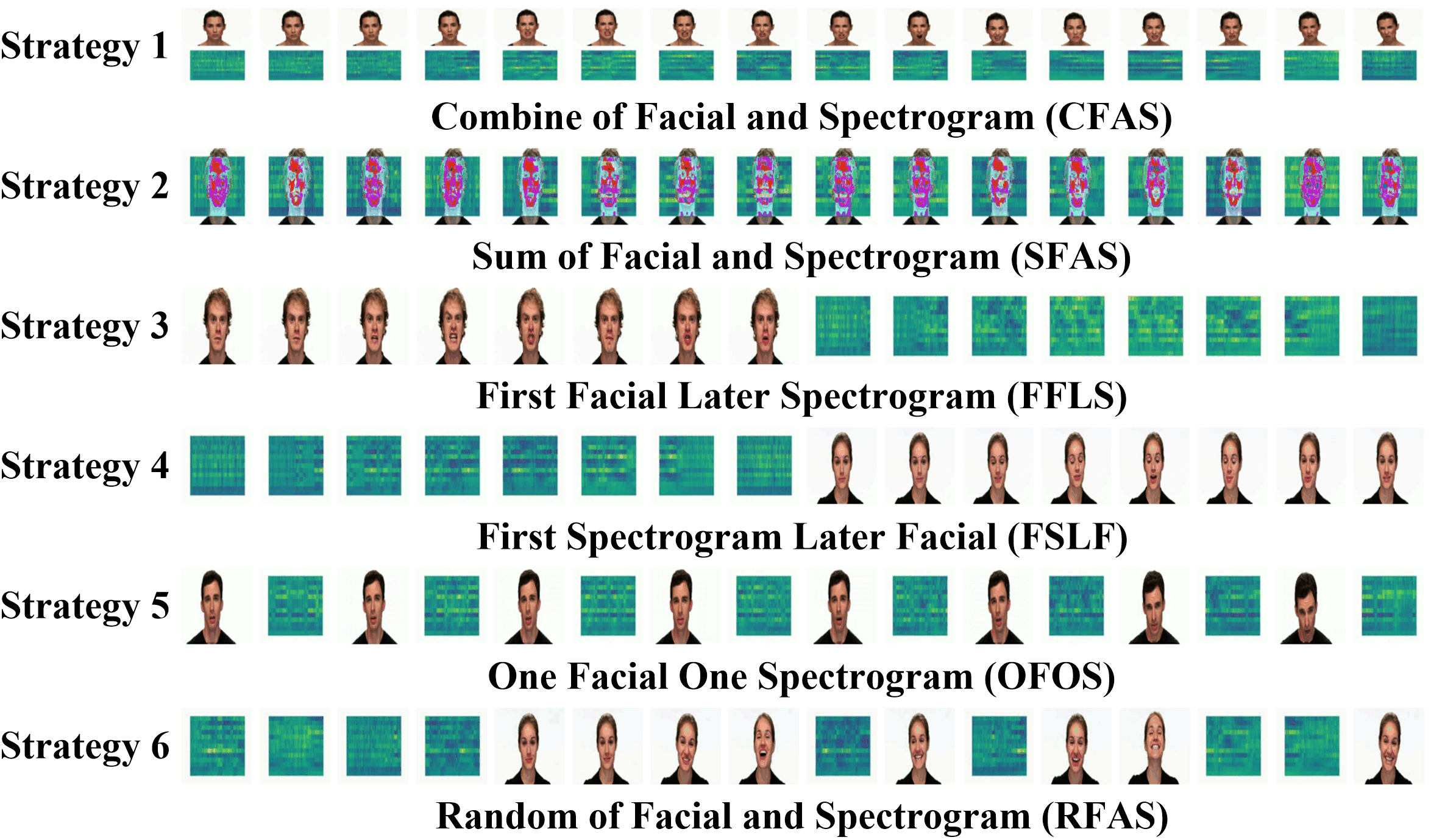}
  \caption{Six Multimodal Sequence Fusion Strategies.}
  \label{figurelabel}
\end{figure}

\section{METHOD}
\subsection{Revisiting VideoMAE}
To illustrate that this work is an inheritance and extension of VideoMAE \cite{c11}, as depicted in Fig. 1, the encoder will directly call the pre-trained model of VideoMAE \cite{c11} encoder as the backbone network. This approach serves to demonstrate effectiveness as a video representation learner while also reducing significant training time. The pre-trained encoder model from VideoMAE \cite{c11} is used, which requires consistency with the input size of $ 16 \times 224 \times 224 $, where 16 is the number of frames, and the image size is $ 224 \times 224 $. Additionally, the patch size remains $2 \times 16 \times 16 $, resulting in 1568 patches as the input size for the token sequence. The VideoMAE \cite{c11} encoder uses ViT-L \cite{c18} as the backbone network with a dimension of 1024. However, this work primarily analyzes multimodal inputs. Therefore, several different sequence fusion methods will be optimized while maintaining the input size and patch size.

\subsection{Multimodal Sequence Strategy}
Firstly, 16 frames of video $ \textbf{V} \in {R}^{16 \times 224 \times 224} $ are uniformly downsampled from the single clip. At the same time, to align with each video frame, the audio signal is uniformly downsampled to generate 16 spectrograms $ \textbf{A} \in {R}^{16 \times 224 \times 224} $. As shown in Fig. 2, in the same time sequence, each sampled time point of the video frame corresponds to the audio signal contained before and after each time point. This ensures the maximum alignment of partial emotion features between the audio signal and facial expression in the video.

On the other hand, in order to verify that extracting spatiotemporally highly correlated dynamic representation information from unified video and audio signals can help improve the accuracy of emotion recognition. We explore the following hypotheses from the sequence fusion strategy of spatiotemporal aspects of audio and video signals:

\begin{itemize}
  \item From the spatial perspective, video signals and audio signals are spliced or superimposed to explore the dynamic representation information of the video and audio signals in spatial integrity under the certain time sequence.
  \item From the temporal perspective, the video signal and audio signal are sequence transform to explore the dynamic representation information of the video and audio signal in time sequence under the condition of spatial integrity.
\end{itemize}

Therefore, as shown in Fig. 3, this work will propose six different multimodal sequence fusion strategies to optimize MultiMAE-DER performance based on the above hypotheses. \textbf{For strategy 1 (CFAS)}, 16 frames of video $ \textbf{V} \in {R}^{16 \times 112 \times 224} $
 and corresponding 16 audio spectrograms $ \textbf{A} \in {R}^{16 \times 112 \times 224} $  will be directly spliced to explore whether there is a certain correlation in the spatiotemporal sequence for this multimodal input $ \textbf{X} \in {R}^{16 \times 224 \times 224} $. \textbf{For strategy 2 (SFAS)}, 16 frames of video $ \textbf{V} \in {R}^{16 \times 224 \times 224} $ and corresponding 16 audio spectrograms $ \textbf{A} \in {R}^{16 \times 224 \times 224} $  will be normalized, and then directly added to verify the feasibility of spatially fusing audio-visual signals. 
 \textbf{For strategy 3 (FFLS)}, considering input size constraints, the single clip are uniformly downsampled to 8 frames of video $ \textbf{V} \in {R}^{8 \times 224 \times 224} $ and corresponding 8 audio spectrograms $ \textbf{A} \in {R}^{8 \times 224 \times 224} $. They are sorted based on facial frames first and then audio spectrograms to consider the impact of solely the temporal sequence in the visual-audio sorting of multimodal input $ \textbf{X} \in {R}^{16 \times 224 \times 224} $ on emotion recognition results. The mathematical expressions for these three multimodal sequence fusion strategies are as follows:

\begin{equation}
\textbf{X}_i = \text{Concat}(\textbf{V}_i, \textbf{A}_i)
\label{eq:1}
\end{equation}
\begin{equation}
\textbf{X}_i = \text{Add}(\text{Norm}(\textbf{V}_i), \text{Norm}(\textbf{A}_i))
\label{eq:2}
\end{equation}
\begin{equation}
\textbf{X} = \text{Seq}(\textbf{V}_1, \textbf{V}_2, \ldots, \textbf{V}_8, \textbf{A}_1, \textbf{A}_2, \ldots, \textbf{A}_8)
\label{eq:sequence}
\end{equation}

\textbf{For strategy 4 (FSLF)}, following the method of Strategy 3, the order of video frames $ \textbf{V} \in {R}^{8 \times 224 \times 224} $ and audio spectrograms $ \textbf{A} \in {R}^{8 \times 224 \times 224} $ are swapped to consider the impact of different orders in the temporal sequence of visual-audio input $ \textbf{X} \in {R}^{16 \times 224 \times 224} $ on emotion recognition results. \textbf{For strategy 5 (OFOS)}, following the order of one video frame $ \textbf{V} \in {R}^{8 \times 224 \times 224}$  followed by one audio spectrogram $ \textbf{A} \in {R}^{8 \times 224 \times 224}$, to explore the impact of different permutations of the time sequence on emotion recognition results.

Finally, \textbf{for strategy 6 (RFAS)}, randomly sorting 8 frames of video $ \textbf{V} \in {R}^{8 \times 224 \times 224}$ and corresponding 8 spectrograms $ \textbf{A} \in {R}^{8 \times 224 \times 224} $ to verify the influence of unordered temporal sequence on visual-audio multimodal data input $ \textbf{X} \in {R}^{16 \times 224 \times 224} $ on emotion recognition results. The mathematical expressions for these three multimodal sequence fusion strategies are as follows:

\begin{equation}
\textbf{X} = \text{Seq}(\textbf{A}_1, \textbf{A}_2, \ldots, \textbf{A}_8, \textbf{V}_1, \textbf{V}_2, \ldots, \textbf{V}_8 )
\label{eq:sequence}
\end{equation}
\begin{equation}
\textbf{X} = \text{Seq}(\textbf{V}_1, \textbf{A}_1, \textbf{V}_2, \textbf{A}_2, \ldots, \textbf{V}_8,\textbf{A}_8 )
\label{eq:sequence}
\end{equation}
\begin{equation}
\textbf{X} = \text{Rand}(\textbf{V}_1, \textbf{V}_2, \ldots, \textbf{V}_8,\textbf{A}_1,\textbf{A}_2, \ldots, \textbf{A}_8)
\label{eq:sequence}
\end{equation}

\subsection{MultiMAE-DER: Model Structure}

As shown in Fig. 1, uniform downsampling is applied to both video and audio in the single clip to align them and meet input size requirements. Subsequently, different sequence fusion strategies from Section III.B are employed as multimodal input $ \textbf{X} \in {R}^{16 \times 224 \times 224}$  to verify the assumptions of this work. Like VideoMAE [11], before entering the encoder, the input data is patched to construct token sequence $ \textbf{S} \in {R}^{1568 \times 512}$. In this model, 3D convolutional kernels are used as embedding layers to perform non-overlapping patching, where the kernel size is $ 2 \times 16 \times 16 $. Linear projection (LP) is applied to each cube patch to flatten the input $\hat{\textbf{X}} \in {R}^{1568 \times 1024}$ into the encoder. After processing through the encoder, the output feature matrix $\textbf{F} \in {R}^{1568 \times 1024}$ with the same input size is obtained. For a given downstream task (dynamic emotion recognition in this case), a single linear layer (MLP) serves as a specific-task head to process the feature matrix $\textbf{F} \in {R}^{1568 \times 1024}$, yielding the final emotion result $\textbf{Y} \in {R}^{1 \times 7}$  for this clip. The mathematical expressions for the above process are as follows:

\begin{equation}
\textbf{S} = \text{Patching}(\textbf{X})
\label{eq:sequence}
\end{equation}
\begin{equation}
\hat{\textbf{X}} = \text{LP}(\textbf{S})
\label{eq:sequence}
\end{equation}
\begin{equation}
\textbf{F} = \text{Encoder}(\hat{\textbf{X}})
\label{eq:sequence}
\end{equation}
\begin{equation}
\textbf{Y} = \text{MLP}({\textbf{F}})
\label{eq:sequence}
\end{equation}

\section{EXPERIMENTS}

\subsection{Datasets} 

\subsubsection{RAVDESS \cite{c19}}  The Ryerson Audio-Visual Database of Emotional Speech and Song is a dataset consisting of emotional performances conducted by 24 North American professional actors in a laboratory environment with North American accent. The emotions depicted in this dataset include anger, disgust, fear, happiness, neutral, sadness, and surprise. Each emotion has two intensity variations: normal and strong. In this work, we utilize only the visual-audio speech dataset, comprising 1440 video files. Among them, there are 288 videos for the neutral emotion, and the remaining emotions each have 192 videos. Our model evaluation is conducted using a 6-fold cross-validation on subjects-independent of the emotion.

\subsubsection{CREMA-D\cite{c20}} Crowd-sourced Emotional Multimodal Actor Dataset comprises emotional performances by 91 professional actors from around the world, representing different countries, and ethnicities. These actors conducted emotional performances in a laboratory environment with diverse accents. Each actor presented six different emotions: anger, disgust, fear, happiness, neutral, and sadness. Additionally, they expressed each emotion at low, medium, high, and unspecified intensity levels to dialogue. The dataset includes a total of 7,442 video clips, with 1,087 clips for the neutral emotion and 1,271 clips for each of the remaining emotions. Model evaluation will be conducted using 5-fold cross-validation on subjects-independent of the emotion.

\begin{figure}[thpb]
  \centering
  \includegraphics[scale=0.345]{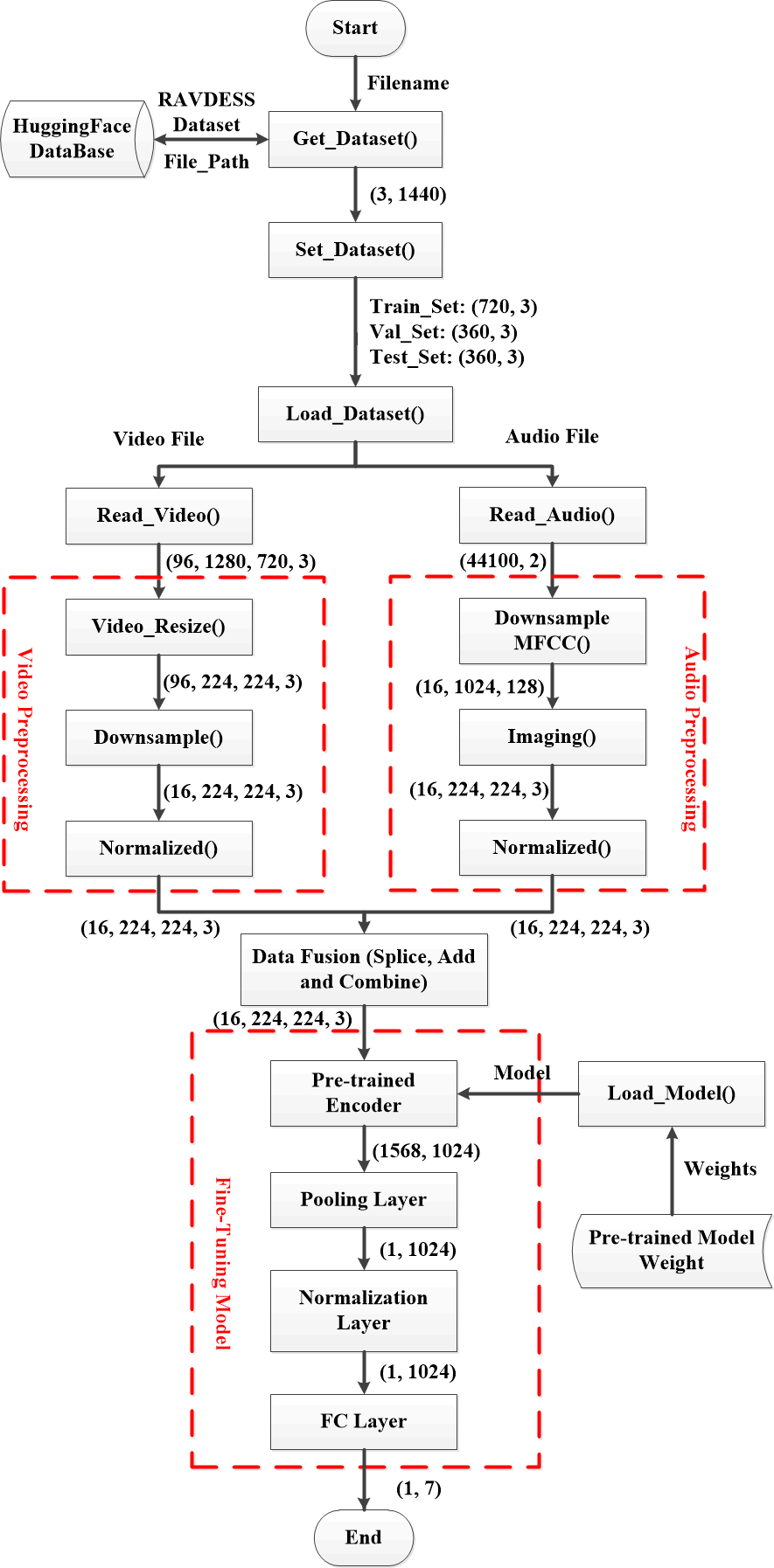}
  \caption{MultiMAE-DER Program Flow-Chart.}
  \label{figurelabel}
\end{figure}

\subsubsection{IEMOCAP\cite{c21}} The Interactive Emotional Dyadic Motion Capture Database is a multimodal and multi-speaker visual-audio dataset featuring improvised emotional binary scene creations by ten non-professional actors. The dataset includes approximately 12 hours of video data, capturing nine different emotional expressions: anger, happiness, excitement, sadness, discouragement, fear, surprise, other, and neutral states. Following the emotion categorization approach of AV-Superb \cite{c22}, we conduct a four-emotion classification task (anger, happiness, sadness, and neutral). To increase the sample size, the data for the excited state are added to the data for the happy state. The fine-tuned model will be evaluated on this dataset using 5-fold cross-validation.

\subsection{Implementation Details}
\subsubsection{Downsampling}
As discussed in Section III.B regarding the strategies for multimodal sequence fusion, downsampling is a crucial step in this proposal. Depending on the different sequence fusion strategies, we have varying uniform downsampling steps. Taking the RAVDESS \cite{c19} dataset as an example, for strategies 1 and 2, where facial videos and audio spectrograms have 16 frames per clip, the downsampling step size is set to 6. For strategies 3, 4, 5, and 6, where facial videos and audio spectrograms have 8 frames per clip, the downsampling step size is set to 12.
The data preprocessing for the above process is illustrated in Fig. 4, depicting video preprocessing and audio preprocessing. For video preprocessing, the cutoff time of a single clip is computed based on the required frame number and downsampling step size, followed by the downsampling of the clip. Simultaneously, for audio preprocessing, the audio signal of the clip is extracted. The time point $T$  corresponding to the facial frame is determined based on the frame number. Then, the audio signal is cut at the time point $T-1$  and $T+1$  to perform Short-Time Fourier Transform (STFT) and Mel-Frequency Cepstral Coefficients (MFCC) to obtain the audio spectrogram corresponding to the facial frame at time point $T$ .

\subsubsection{Pre-training}
After preprocessing, the obtained facial frames and audio spectrograms implement sequence fusion, which includes methods such as splice, addition, and combine. For the pre-trained encoder model, to eliminate domain differences, the pre-trained encoder model used in our proposal is derived from MAE-DFER \cite{c14}. This encoder is built exclusively upon VideoMAE \cite{c11} and has undergone self-supervised training on the extensive dataset of facial videos from YouTube, aiming to mitigate domain differences.

\subsubsection{Fine-tuning}
As described in Section III.C, to minimize the interference of inductive bias, our approach utilizes only single-layer neural network as a specific-task head for fine-tuning on the downstream task. The fine-tuning employs the AdamW \cite{c23} optimizer with the base learning rate of 1e-2, weight decay of 0.05, sparse categorical cross-entropy as the loss function, and sparse categorical accuracy as the metric function. The evaluation metrics used by the model include unweighted average recall (UAR) and weighted average recall (WAR).

\subsection{Results}

In this section, we first compared MultiMAE-DER with other previous state-of-the-art multimodal models on the RAVDESS \cite{c19}. As shown in Table 1, in the multimodal model domain, the results exhibit significant variations based on different learning methods, whether supervised learning or self-supervised learning. The self-supervised multimodal model VQ-MAE-AV \cite{c30} outperforms the supervised multimodal model AVT \cite{c29} by 4\% (83.20\% vs. 79.20\%) in terms of WAR. Furthermore, our MultiMAE-DER model, employing different multimodal sequence fusion strategies, generally surpasses the supervised multimodal model. Particularly, when using strategy 4, MultiMAE-DER-FSLF outperforms the supervised model by 4.41\% (83.61\% vs. 79.20\%) in terms of WAR. Among the self-supervised multimodal models, our optimal strategy (i.e., strategy 4 FSLF) of MultiMAE-DER also outperforms the multimodal model VQ-MAE-AV \cite{c30} by 0.41\% (83.61\% vs. 83.20\%) in terms of WAR.

\begin{table}
\caption{Results on RAVDESS (7-class). SSL: self-supervised learning method or not. UAR: unweighted average recall. WAR: weighted average recall. Underlined: the best supervised result. Bold: the best result.}
\label{table_example}
\begin{center}
\begin{tabular}{lcccc}
\hline
Method & SSL & Modality & UAR & WAR\\
\hline
AV-LSTM\cite{c15} & $\chi$ & V+A & $-$ & 65.80\\
AV-Gating\cite{c15} & $\chi$ & V+A & $-$ & 67.70 \\
MCBP\cite{c24} & $\chi$ & V+A & $-$ & 71.32 \\
MMTM\cite{c25} & $\chi$ & V+A & $-$ & 73.12 \\
ERANNs\cite{c26} & $\chi$ & V+A & $-$ & 74.80 \\
MSAF\cite{c16} & $\chi$ & V+A & $-$ & 74.86 \\
SFN-SR\cite{c17} & $\chi$ & V+A & $-$ & 75.76 \\
MATER \cite{c27} & $\chi$ & V+A & $-$ & 76.30 \\
MulT \cite{c28} & $\chi$ & V+A & $-$ & 76.60 \\
AVT \cite{c29} & $\chi$ & V+A & $-$ & \underline{79.20} \\
VQ-MAE-AV \cite{c30} & $\checkmark$ & V+A & $-$ & 83.20 \\
\hline
MultiMAE-DER & $\checkmark$ & V & $-$ & 74.13 \\
MultiMAE-DER & $\checkmark$ & A & $-$ & 80.55 \\
MultiMAE-DER-RFAS & $\checkmark$ & V+A & $75.97$ & 75.44 \\
MultiMAE-DER-SFAS & $\checkmark$ & V+A & $75.79$ & 76.94 \\
MultiMAE-DER-OFOS & $\checkmark$ & V+A & $77.78$ & 78.61 \\
MultiMAE-DER-CFAS & $\checkmark$ & V+A & $80.65$ & 81.39 \\
MultiMAE-DER-FFLS & $\checkmark$ & V+A & $82.27$ & 83.56 \\
MultiMAE-DER-FSLF & $\checkmark$ & V+A & $\textbf{83.23}$ & $\textbf{83.61}$ \\
\hline
\end{tabular}
\end{center}
\end{table}

The visual-audio results on the CREMA-D \cite{c20} are presented in Table 2. We observe that the optimal strategy of MultiMAE-DER achieves the WAR result surpassing the VQ-MAE-AV \cite{c30} multimodal self-supervised learning model by 0.96\% (79.36\% vs. 78.40\%). Compared to supervised learning approaches, the optimal strategy of MultiMAE-DER outperforms the multimodal model RAVER \cite{c34} by 2.06\% (79.36\% vs. 77.30\%) in WAR. In addition, most of the results of other strategies in MultiMAE-DER outperform state-of-the-art multimodal supervised learning models. This indicates the effectiveness of multimodal sequence fusion strategies, which can aid in identifying spatio-temporal correlations across cross-domain data to some extent.

\begin{table}
\caption{Results on CREMA-D (6-class). SSL: self-supervised learning method or not. UAR: unweighted average recall. WAR: weighted average recall. Underlined: the best supervised result. Bold: the best result.}
\label{table_example}
\begin{center}
\begin{tabular}{lcccc}
\hline
Method & SSL & Modality & UAR & WAR\\
\hline
EF-GRU \cite{c31} & $\chi$ & V+A & $-$ & 57.06\\
LF-GRU \cite{c31} & $\chi$ & V+A & $-$ & 58.53 \\
TFN \cite{c32} & $\chi$ & V+A & $-$ & 63.09 \\
MATER \cite{c27} & $\chi$ & V+A & $-$ & 67.20 \\
AuxFormer\cite{c33} & $\chi$ & V+A & $-$ & 71.70 \\
AV-LSTM\cite{c15} & $\chi$ & V+A & $-$ & 72.90 \\
AV-Gating\cite{c15} & $\chi$ & V+A & $-$ & 74.00 \\
RAVER \cite{c34} & $\chi$ & V+A & $-$ & \underline{77.30} \\
VQ-MAE-AV \cite{c30} & $\checkmark$ & V+A & $-$ & 78.40 \\
\hline
MultiMAE-DER & $\checkmark$ & V & $-$ & 77.83 \\
MultiMAE-DER & $\checkmark$ & A & $-$ & 78.45 \\
MultiMAE-DER-RFAS & $\checkmark$ & V+A & $74.62$ & 74.90 \\
MultiMAE-DER-SFAS & $\checkmark$ & V+A & $75.73$ & 75.48 \\
MultiMAE-DER-OFOS & $\checkmark$ & V+A & $76.88$ & 76.54 \\
MultiMAE-DER-CFAS & $\checkmark$ & V+A & $78.24$ & 78.16 \\
MultiMAE-DER-FFLS & $\checkmark$ & V+A & $78.59$ & 78.83 \\
MultiMAE-DER-FSLF & $\checkmark$ & V+A & $\textbf{79.12}$ & $\textbf{79.36}$ \\
\hline
\end{tabular}
\end{center}
\end{table}

In Table 3, we compare state-of-the-art self-supervised multimodal models on the IEMOCAP \cite{c21}. The results are derived from the evaluation test conducted by AV-Superb \cite{c22}. The results showed that in the field of self-supervised learning, our MultiMAE-DER optimal strategy model (FSLF) outperforms the multimodal model AVBERT \cite{c37} by 1.86\% (63.73\% vs. 61.87\%) in WAR. Upon observing the results of multiple strategies in our model and comparing them with the results of these three self-supervised multimodal models, we notice that MultiMAE-DER model performs slightly better in handling cross-domain data. This indirectly reflects the model's ability to extract representational information regarding spatio-temporal correlations across cross-domain data. Moreover, these dynamic features play a crucial role in inferring emotional context.

\begin{table}
\caption{Results on IEMOCAP (4-class). SSL: self-supervised learning method or not. UAR: unweighted average recall. WAR: weighted average recall. Bold: the best result.}
\label{table_example}
\begin{center}
\begin{tabular}{lcccc}
\hline
Method & SSL & Modality & UAR & WAR\\
\hline
AV-HuBERT\cite{c35} & $\checkmark$ & V+A & $-$ & 46.45 \\
MAViL\cite{c36} & $\checkmark$ & V+A & $-$ & 54.94 \\
AVBERT \cite{c37} & $\checkmark$ & V+A & $-$ & 61.87 \\
\hline
MultiMAE-DER & $\checkmark$ & V & $-$ & 56.13 \\
MultiMAE-DER & $\checkmark$ & A & $-$ & 58.69 \\
MultiMAE-DER-RFAS & $\checkmark$ & V+A & $58.62$ & 59.98 \\
MultiMAE-DER-SFAS & $\checkmark$ & V+A & $60.39$ & 60.17 \\
MultiMAE-DER-OFOS & $\checkmark$ & V+A & $61.87$ & 61.12 \\
MultiMAE-DER-CFAS & $\checkmark$ & V+A & $61.98$ & 62.25 \\
MultiMAE-DER-FFLS & $\checkmark$ & V+A & $62.92$ & 63.43 \\
MultiMAE-DER-FSLF & $\checkmark$ & V+A & $\textbf{63.21}$ & $\textbf{63.73}$ \\
\hline
\end{tabular}
\end{center}
\end{table}

In addition, compared with the MultiMAE-DER single-modal model, the MultiMAE-DER multi-modal optimal strategy model (FSLF) is far superior in performance to the visual-only model and the audio-only model. The results show that the WAR on RAVDESS \cite{c19} is enhanced by 9.48\% (83.61\% vs. 74.13\%) and 3.06\% (83.61\% vs. 80.55\%) respectively, the WAR on CREMA-D \cite{c20} is enhanced by 1.53\% (79.36\% vs. 77.83\%) and 0.91\% (79.36\% vs. 78.45\%) respectively, and the WAR on IEMOCAP \cite{c21} is enhanced by 7.60\% (63.73\% vs. 56.13\%) and 5.04\% (63.73\% vs. 58.69\%) respectively.

Upon analyzing the outcomes across the three datasets, we observed that MultiMAE-DER exhibits negligible discrepancies in emotion recognition results between strategy 3 (FFLS) and 4 (FSLF). This aligns with expectations, given that their distinctions lie solely in the sorting of multimodal sequences. Furthermore, the effects of these two strategies surpass those of other strategies, demonstrating enhancements ranging from 2\% to 8\% in WAR on the RAVDESS \cite{c19}, 1\% to 5\% on the CREMA-D \cite{c20}, and 1\% to 3\% on IEMOCAP \cite{c21}.

According to our analysis, different multimodal sequence fusion strategies determine the state of the token sequence. As we know, in the ViT \cite{c18}, the sorting of the token sequence is constructed by the linear projection of patches. In our strategy, for strategy 1, the top half of all frames consists of visual data, while the bottom half consists of audio data. After patching, every 98 sequences among the 1568 token sequences will undergo cross-domain sequence exchange. Strategy 2 is the sum of cross-domain data normalization, which will directly disrupt the overall spatial structure. Strategy 3, which involves sorting visual data and the corresponding audio data, results in 1568 token sequences where the first 784 sequences are visual, and the latter 784 sequences are audio. Strategy 4 involves swapping the order of visual data and audio data from strategy 3. In the resulting 1568 token sequences, the first 784 sequences are audio, and the latter 784 sequences are visual. Strategy 5 involves interleaving the order of visual data and corresponding audio data. In the resulting 1568 token sequences, half of each sequence's dimensions are visual features, while the other half consists of audio features. Strategy 6 involves randomly sorting visual data and audio data. The resulting 1568 token sequences lack coherence, whether visual-audio sequence or spatio-temporal sequence. Finally, we can draw the following observations:

\begin{itemize}
  \item Strategy 1 exhibits temporal continuity but lacks spatial continuity.
  \item Strategy 2 has temporal continuity but disrupts the overall spatial sequence structure.
  \item Strategies 3 and 4 demonstrate both spatial and temporal continuity, with a high concentration of spatio-temporal correlation.
  \item Strategy 5 shows spatial continuity but disrupts the overall temporal sequence structure.
  \item Strategy 6 lacks both spatial and temporal continuity, simultaneously disrupting the overall spatio-temporal correlation.
\end{itemize}

\section{CONCLUSIONS AND FUTURE WORK}
\subsection{CONCLUSIONS}
This work represents a new exploration of the approach to handling multimodal inputs, establishing a novel framework for the integrated processing of multimodal data under the conditions of self-supervised learning architecture. This paper investigates six different multimodal sequence fusion strategies to explore the diverse interpretations of multimodal representation information extracted by the same model when exposed to different spatio-temporal sequence inputs. Results indicate that fusing multimodal data on spatio-temporal sequences significantly improves the model performance by capturing correlations between cross-domain data. Furthermore, MultiMAE-DER validates the performance effectiveness of self-supervised learning as an efficient learner, contributing to the potential of the masked autoencoder as a unified framework for solving contextual semantic inference problems.

\subsection{FUTURE WORK}
In future work, the framework can be further developed from two aspects. In terms of depth, the self-supervised learning approach of the multimodal masked autoencoder involves reconstructing the input of multimodal visual-audio data to enhance the pre-training encoder. This enhanced model can subsequently function as a multimodal pre-trained model, substituting the video autoencoder pre-trained model utilized in this study. In terms of breadth, this work has covered the aspects of ``see" (video) and ``listen" (audio) in the model feature extraction process for input data, which is essential for further incorporating ``read" (text) into the model inference process.


\begin{thebibliography}{00}
\bibitem{c1} Arriaga, Octavio, Matias Valdenegro-Toro, and Paul Plöger. "Real-time convolutional neural networks for emotion and gender classification." arXiv preprint arXiv:1710.07557 (2017).
\bibitem{c2} Minaee, Shervin, Mehdi Minaei, and Amirali Abdolrashidi. "Deep-emotion: Facial expression recognition using attentional convolutional network." Sensors 21.9 (2021): 3046.
\bibitem{c3} Ngwe, Jia Le, et al. "PAtt-Lite: Lightweight Patch and Attention MobileNet for Challenging Facial Expression Recognition." arXiv preprint arXiv:2306.09626 (2023).
\bibitem{c4} Zhang, Yongqiang, et al. "Dynamic gesture recognition model based on millimeter-wave radar with ResNet-18 and LSTM." Frontiers in Neurorobotics 16 (2022): 903197.
\bibitem{c5} Ouyang, Xi, et al. "A 3D-CNN and LSTM based multi-task learning architecture for action recognition." IEEE Access 7 (2019): 40757-40770.
\bibitem{c6} Vaswani, Ashish, et al. "Attention is all you need." Advances in neural information processing systems 30 (2017).
\bibitem{c7} Zhao, Zengqun, and Qingshan Liu. "Former-dfer: Dynamic facial expression recognition transformer." Proceedings of the 29th ACM International Conference on Multimedia. 2021.
\bibitem{c8} Ma, Fuyan, Bin Sun, and Shutao Li. "Spatio-temporal transformer for dynamic facial expression recognition in the wild." arXiv preprint arXiv:2205.04749 (2022).
\bibitem{c9} Devlin, Jacob, et al. "Bert: Pre-training of deep bidirectional transformers for language understanding." arXiv preprint arXiv:1810.04805 (2018).
\bibitem{c10} He, Kaiming, et al. "Masked autoencoders are scalable vision learners." Proceedings of the IEEE/CVF conference on computer vision and pattern recognition. 2022.
\bibitem{c11} Tong, Zhan, et al. "Videomae: Masked autoencoders are data-efficient learners for self-supervised video pre-training." Advances in neural information processing systems 35 (2022): 10078-10093.
\bibitem{c12} Huang, Po-Yao, et al. "Masked autoencoders that listen." Advances in Neural Information Processing Systems 35 (2022): 28708-28720.
\bibitem{c13} Hsu, Wei-Ning, et al. "Hubert: Self-supervised speech representation learning by masked prediction of hidden units." IEEE/ACM Transactions on Audio, Speech, and Language Processing 29 (2021): 3451-3460.
\bibitem{c14} Sun, Licai, et al. "Mae-dfer: Efficient masked autoencoder for self-supervised dynamic facial expression recognition." Proceedings of the 31st ACM International Conference on Multimedia. 2023.
\bibitem{c15} Ghaleb, Esam, Mirela Popa, and Stylianos Asteriadis. "Multimodal and temporal perception of audio-visual cues for emotion recognition." 2019 8th International Conference on Affective Computing and Intelligent Interaction (ACII). IEEE, 2019.
\bibitem{c16} Su, Lang, et al. "Msaf: Multimodal split attention fusion." arXiv preprint arXiv:2012.07175 (2020).
\bibitem{c17} Fu, Ziwang, et al. "A cross-modal fusion network based on self-attention and residual structure for multimodal emotion recognition." arXiv preprint arXiv:2111.02172 (2021).
\bibitem{c18} Dosovitskiy, Alexey, et al. "An image is worth 16x16 words: Transformers for image recognition at scale." arXiv preprint arXiv:2010.11929 (2020).
\bibitem{c19} Livingstone, Steven R., and Frank A. Russo. "The Ryerson Audio-Visual Database of Emotional Speech and Song (RAVDESS): A dynamic, multimodal set of facial and vocal expressions in North American English." PloS one 13.5 (2018): e0196391.
\bibitem{c20} Cao, Houwei, et al. "Crema-d: Crowd-sourced emotional multimodal actors dataset." IEEE transactions on affective computing 5.4 (2014): 377-390.
\bibitem{c21} Busso, Carlos, et al. "IEMOCAP: Interactive emotional dyadic motion capture database." Language resources and evaluation 42 (2008): 335-359.
\bibitem{c22} Tseng, Yuan, et al. "Av-superb: A multi-task evaluation benchmark for audio-visual representation models." arXiv preprint arXiv:2309.10787 (2023).
\bibitem{c23} Loshchilov, Ilya, and Frank Hutter. "Decoupled weight decay regularization." arXiv preprint arXiv:1711.05101 (2017).
\bibitem{c24} Fukui, Akira, et al. "Multimodal compact bilinear pooling for visual question answering and visual grounding." arXiv preprint arXiv:1606.01847 (2016).
\bibitem{c25} Joze, Hamid Reza Vaezi, et al. "MMTM: Multimodal transfer module for CNN fusion." Proceedings of the IEEE/CVF conference on computer vision and pattern recognition. 2020.
\bibitem{c26} Verbitskiy, Sergey, Vladimir Berikov, and Viacheslav Vyshegorodtsev. "Eranns: Efficient residual audio neural networks for audio pattern recognition." Pattern Recognition Letters 161 (2022): 38-44.
\bibitem{c27} Ghaleb, Esam, Jan Niehues, and Stylianos Asteriadis. "Multimodal attention-mechanism for temporal emotion recognition." 2020 IEEE International Conference on Image Processing (ICIP). IEEE, 2020.
\bibitem{c28} Tsai, Yao-Hung Hubert, et al. "Multimodal transformer for unaligned multimodal language sequences." Proceedings of the conference. Association for computational linguistics. Meeting. Vol. 2019. NIH Public Access, 2019.
\bibitem{c29} Chumachenko, Kateryna, Alexandros Iosifidis, and Moncef Gabbouj. "Self-attention fusion for audiovisual emotion recognition with incomplete data." 2022 26th International Conference on Pattern Recognition (ICPR). IEEE, 2022.
\bibitem{c30} Sadok, Samir, Simon Leglaive, and Renaud Séguier. "A vector quantized masked autoencoder for speech emotion recognition." 2023 IEEE International conference on acoustics, speech, and signal processing workshops (ICASSPW). IEEE, 2023.
\bibitem{c31} Tran, Minh, and Mohammad Soleymani. "A pre-trained audio-visual transformer for emotion recognition." ICASSP 2022-2022 IEEE International Conference on Acoustics, Speech and Signal Processing (ICASSP). IEEE, 2022.
\bibitem{c32} Zadeh, Amir, et al. "Tensor fusion network for multimodal sentiment analysis." arXiv preprint arXiv:1707.07250 (2017).
\bibitem{c33} Goncalves, Lucas, and Carlos Busso. "AuxFormer: Robust approach to audiovisual emotion recognition." ICASSP 2022-2022 IEEE International Conference on Acoustics, Speech and Signal Processing (ICASSP). IEEE, 2022.
\bibitem{c34} Goncalves, Lucas, and Carlos Busso. "Robust audiovisual emotion recognition: Aligning modalities, capturing temporal information, and handling missing features." IEEE Transactions on Affective Computing 13.4 (2022): 2156-2170.
\bibitem{c35} Shi, Bowen, et al. "Learning audio-visual speech representation by masked multimodal cluster prediction." arXiv preprint arXiv:2201.02184 (2022).
\bibitem{c36} Huang, Po-Yao, et al. "Mavil: Masked audio-video learners." Advances in Neural Information Processing Systems 36 (2024).
\bibitem{c37} Lee, Sangho, et al. "Parameter efficient multimodal transformers for video representation learning." arXiv preprint arXiv:2012.04124 (2020).

\end{thebibliography}
\end{document}